\documentclass[sigconf]{acmart}

\usepackage{booktabs} 
\usepackage{subcaption}
\usepackage{multirow}
\usepackage{enumitem}
\usepackage{array}
\usepackage{bm}
\usepackage{amsthm}
\usepackage{amsmath}
\usepackage{color}
\newtheorem{definition}{Definition}

\usepackage{threeparttable}
\usepackage[normalem]{ulem}
\useunder{\uline}{\ul}{}
\newcolumntype{L}[1]{>{\raggedright\let\newline\\\arraybackslash\hspace{0pt}}m{#1}}
\newcolumntype{C}[1]{>{\centering\let\newline\\\arraybackslash\hspace{0pt}}m{#1}}
\newcolumntype{R}[1]{>{\raggedleft\let\newline\\\arraybackslash\hspace{0pt}}m{#1}}

\usepackage[title]{appendix}
\usepackage[linesnumbered,ruled,lined]{algorithm2e}

\setcopyright{none}

\begin{document}
\title{Multi-Aspect Temporal Network Embedding: \\ A Mixture of Hawkes Process View}


\author{Yutian Chang}
\affiliation{
	\department{School of Economics and Management}
	\institution{Beihang University}
	\city{Beijing 100191, China}
}
\email{ytchang@buaa.edu.cn}

\author{Guannan Liu}\authornote{Corresponding author}
\affiliation{
	\department{School of Economics and Management}
	\institution{Beihang University}
	\city{Beijing 100191, China}
}
\email{liugn@buaa.edu.cn}

\author{Yuan Zuo}
\affiliation{
	\department{School of Economics and Management}
	\institution{Beihang University}
	\city{Beijing 100191, China}
}
\email{zuoyuan@buaa.edu.cn}

%
%
%
%
\author{Junjie Wu}
\affiliation{
	\department[0]{School of Economics and Management}
	\department[1]{Beijing Advanced Innovation Center for Big Data and Brain Computing}
	\institution{Beihang University}
	\city{Beijing 100191, China}
}
\additionalaffiliation{
	\department{Beijing Key Laboratory of Emergency Support Simulation Technologies for City Operations, }
	\institution{Beihang University}
	\city{Beijing 100191, China}
}
\email{wujj@buaa.edu.cn}

\begin{abstract}
Recent years have witnessed the tremendous research interests in network embedding. Extant works have taken the neighborhood formation as the critical information to reveal the inherent dynamics of network structures, and suggested encoding temporal edge formation sequences to capture the historical influences of neighbors. In this paper, however, we argue that the edge formation can be attributed to a variety of driving factors including the temporal influence, which is better referred to as multiple \emph{aspects}. As a matter of fact, different node aspects can drive the formation of distinctive neighbors, giving birth to the \emph{multi-aspect embedding} that relates to but goes beyond a temporal scope. Along this vein, we propose a Mixture of Hawkes-based Temporal Network Embeddings (MHNE) model to capture the aspect-driven neighborhood formation of networks. In MHNE, we encode the multi-aspect embeddings into the mixture of Hawkes processes to gain the advantages in modeling the excitation effects and the latent aspects. Specifically, a graph attention mechanism is used to assign different weights to account for the excitation effects of history events, while a Gumbel-Softmax is plugged in to derive the distribution over the aspects. Extensive experiments on 8 different temporal networks have demonstrated the great performance of the multi-aspect embeddings obtained by MHNE in comparison with the state-of-the-art methods.
\end{abstract}


\ccsdesc[500]{Information systems~Data mining}
\ccsdesc[300]{Information systems~Network data models}
\ccsdesc[500]{Computing methodologies~Dimensionality reduction and manifold learning}

\keywords{Network embedding, Hawkes process, Multi-aspect, Temporal network}

\maketitle

\section{Introduction}
Network embedding has attracted extensive attentions in recent years due to its efficiency in encoding network structure into low dimensional representations~\cite{tang2015line,grover2016node2vec,perozzi2014deepwalk,hamilton2017inductive,dai2018adversarial}, which can be flexibly applied to various downstream tasks. In addition to embedding static networks, some efforts have also been made to tackle the dynamic evolving structure of temporal networks~\cite{goyal2018dyngem,pareja2020evolvegcn,deng2019learning,zuo2018embedding,lu2019temporal}. Conventional temporal network embedding methods~\cite{goyal2018dyngem,pareja2020evolvegcn} model the network dynamics via multiple snapshots derived from different time periods, which can only represent the network structure in a certain time window. To overcome the limitations of the snapshot-based methods, recent work~\cite{zuo2018embedding,lu2019temporal} has attempted to track the \emph{neighborhood formation sequence} of each node, where the connection between a target node and the source node primarily depends on the similarity between the two nodes and the influence of the historical target nodes. In this regard, a complete temporal process of network formation can be revealed and hence the derived embeddings can reflect the evolution patterns of nodes.

\begin{figure}[t]
	\centering
	\includegraphics[width=0.5\textwidth]{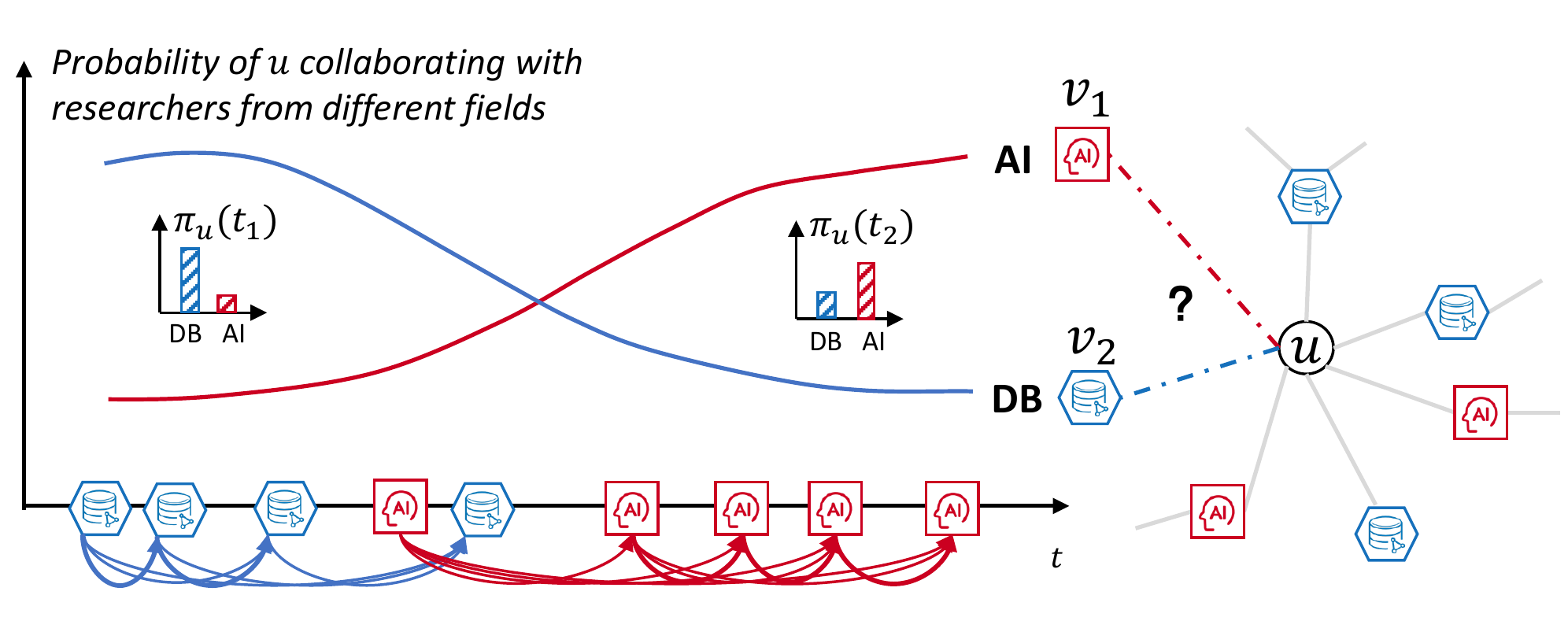}
	\vspace{-0.3cm}
	\caption{An example of considering the history sequence and multiple aspects of nodes}
	\vspace{-0.1cm}
	\label{Fig: Toy Example}
\end{figure}

However, except for the temporal view of the neighborhood formation, as a matter of fact, the connections between a source node and its target nodes may be driven by different underlying factors, which can be referred to as \emph{aspects}. The aspects of nodes can have different meanings in terms of different networks. Especially, the aspects may change over time, and therefore can induce distinctive neighborhood formation of a node. For example, assume that node $u$ in Fig.~\ref{Fig: Toy Example} is a researcher who focused on \emph{Database}~(DB) at the early stages and then shifts his interest to \emph{Artificial Intelligence}~(AI). The left part of the Fig.~\ref{Fig: Toy Example} illustrates the historical coauthors of $u$ listed chronologically on the x-axis, where the y-axis denotes the probability of node $u$ collaborating with researchers from different fields~({\it i.e.}, aspects). The blue/red line in the figure represents the activeness of $u$ in terms of DB/AI, as illustrated, the blue line has a clear downward trend while the red one keeps rising over time. Therefore, given two nodes $v_1$ and $v_2$ representing researchers who are currently focusing on AI and DB respectively, we can easily infer that $u$ has a larger chance of establishing co-authorship with $v_1$ than $v_2$ according to $u$'s neighborhood sequence. In contrast, from a static point of view as shown on the rightmost part of Fig.~\ref{Fig: Toy Example}, where the neighbors of $u$ are regard as an unordered static node set, future connections of $u$ are much less predictable. As seen from the example, the formation of temporal edges are indeed driven by different aspects of nodes, which is essential to understand the underlying mechanism of network evolution. However, prior temporal network embedding methods generally derive a single embedding for each node, and merely address different node aspects.

Given that nodes in networks may have multiple aspects, several prior work~\cite{epasto2019single,liu2019single,park2020unsupervised,wang2019mcne,yang2018multi} has attempted to represent each node with multiple embeddings. Compared to a single embedding vector, multiple embeddings enlarges the embedding space, with each embedding exclusively focusing on one specific aspect. However, these multi-aspect embeddings are derived from a static network, which indeed has some inherent limitations. On the one hand, the neighborhood of a node is generally modeled as an unordered node set, where neighbors that formed at distant timestamps would not be treated differently, thus the derived multi-aspect embedding  may simply be an aggregated distribution over the aspects of all the neighbors.
On the other, the node aspects can change over time, and there might be a focal aspect at one time but the static view fails to capture this. Thus, it cannot precisely predict the future neighbor formation with the out-dated aspect information. Apparently, the temporal network setting is more desirable for multi-aspect embeddings. As a matter of fact, temporal network embedding and the multi-aspect embedding can indeed be regarded as the two sides of a coin and should complement with each other. As far as we are concerned, we are among the first to take a temporal view to derive the multi-aspect embeddings.

To tackle the above-mentioned challenges, we extend the \emph{neighborhood formation} to a notion of \textit{aspect driven edge formation} by plugging an aspect distribution for each edge formed at specific time to account for the aspect-drive factors, which gives rise to a \textit{Multi-Aspect Temporal Network}. In order to obtain the multi-aspect embedding from the temporal network, we propose an end-to-end method named {\it \textbf{M}ixture of \textbf{H}awkes-based Temporal \textbf{N}etwork \textbf{E}mbedding} (MHNE) in this paper. In overall, we model the aspect-driven neighborhood formation with a mixture of Hawkes process (MHP)~\cite{yang2013mixture}, which has great merits in capturing both the temporal excitation effects of historical events and the latent aspects within the sequence. In particular, we exploit the MHP to model the edge formation probability by feeding the multi-aspect embeddings into the intensity function. More specifically, the aspect distribution is induced via Gumbel-Softmax~\cite{jang2016categorical} with a trainable temperature parameter that controls the extent to which we desire a one-hot alike distribution. Additionally, graph attention mechanism~\cite{velivckovic2017graph} is incorporated to assign different weights to the excitation effects of history events. We have conducted extensive experiments on eight different real-world networks by utilizing the the multi-aspect embeddings for several downstream including link prediction, temporal node recommendation. The experimental results have demonstrated the superiority of MHNE over the state-of-the-art methods, and we have also rigorously validated the effectiveness the modeling components, especially the quality of aspects.

\section{Preliminary}
\subsection{Aspect Driven Neighborhood Formation}
Recent works~\cite{zhou2018dynamic,lu2019temporal} try to learn temporal network embedding by looking into the formation process of the network, specifically, the neighborhood formation process of nodes. However, as illustrated in the toy example, they ignore a fact that the evolution of edges are inherently driven by the underlying aspects, which provides a more comprehensive view of the neighborhood formation process. Based on the above consideration, we formally define the multi-aspect temporal network.

\begin{definition}[Multi-Aspect Temporal Network]
	Multi-aspect temporal network is a network with aspects driven temporal edges, which can be denoted as $\mathcal{G}=<\mathcal{V}, \mathcal{E}; \mathcal{H}, \mathcal{K}>$, where $\mathcal{V}$ denotes the set of nodes, $\mathcal{E}$ denotes the set of temporal edges, $\mathcal{H}$ denotes the set of neighborhood formation sequences and $\mathcal{K}$ denotes the set of aspect distributions. Each temporal edge $e=(u,v,t) \in \mathcal{E}$ between the source node $u$ and the target node $v$ at time $t$ is driven by the aspect distribution $\mathcal{K}_u(t)$ and the neighborhood formation sequence $\mathcal{H}_u(t)=\{(h, t_h)|(u, h, t_h)\in\mathcal{E}, t_h<t\}$ of the source node $u$.
	\label{def:matn}
\end{definition}

From the above definition, we can find the evolution of edges in the multi-aspect temporal network is still modeled with neighborhood formation sequence of the source node. The difference is the evolution of edges is also driven by the unobserved aspect distribution of the source node, with which a much clearer clue is provided for predicting future connections of the source node. Therefore, we formally define the aspect driven edge formation.

\begin{definition}[Aspect Driven Edge Formation]
	\label{def:def_2}
	Given a temporal edge $e = (u, v, t)$ and the neighborhood formation sequence $\mathcal{H}_u(t)$ of its source node $u$ at time $t$, the formation probability of $e$, in other words, the probability of $u$ connecting to $v$ at time $t$ is $p(v|u,t,\mathcal{H}_u(t))$. Given the aspect distribution $\mathcal{K}_u(t)$ of the source node $u$ at time $t$, the formation probability of temporal edge $e$ driven by the aspects can be decomposed as:
	\begin{equation}
	\label{eq:form_prob}
	p(v|u,t,\mathcal{H}_u(t)) = \sum_{k=1}^{|\mathcal{K}_u(t)|} \mathcal{K}_u^k(t) p(v|u,t,\mathcal{H}_u(t),k),
	\end{equation}
	where $\mathcal{K}_u^k(t)=p(k|u,t)$ is the probability of $k$-th aspect of the source node $u$ at time $t$, $p(v|u,t,\mathcal{H}_u(t),k)$ is the formation probability of $e$ driven by the neighborhood formation sequence $\mathcal{H}_u(t)$ and the $k$-th aspect.
\end{definition}

With the aspect driven edge formation defined above, we can easily understand how the neighborhood formation sequence $\mathcal{H}_u(t)$ of the source node $u$ is driven by its aspects distribution $\mathcal{K}_{u}(t)$. That is, for each pair of history target node $h$ and timestamp $h_t$ in $\mathcal{H}_u(t)$, there is an edge $(u, h, t_h)$ that is driven by $\mathcal{K}_{u}(t)$. Once all nodes' neighborhood formation sequences at all timestamps are driven by the aspects, the entire multi-aspect temporal network will be established.

\subsection{Problem Definition}
Given a multi-aspect temporal network $\mathcal{G}=<\mathcal{V}, \mathcal{E}; \mathcal{H}, \mathcal{K}>$, the neighborhood formation $\mathcal{H}_u(t)$ of each node $u \in \mathcal{V}$ can be induced by tracking all the timestamps when $u$ interacts with its neighbors before time $t$. However, the aspect distribution $\mathcal{K}_u(t)$ of each node $u$ at time $t$ is unobserved. To learn multiple aspects that driven the entire temporal network, we resort to learn the multi-aspect temporal network embedding as defined in the follow.

\begin{definition}[Multi-Aspect Temporal Network Embedding]
	Given the multi-aspect temporal network $\mathcal{G}=<\mathcal{V}, \mathcal{E}; \mathcal{H}, \mathcal{K}>$, we aim to learn an embedding function $\phi: u \in \mathcal{V}\rightarrow\{\mathbf{I}_u\in\mathbb{R}^m, \mathbf{A}_u\in\mathbb{R}^{K*m}\}$, where $\mathbf{I}_u$ is the identity embedding of node $u$, and $\mathbf{A}_u^k$ represents the embedding of $k$-th aspect of node $u$, $K$ is the number of aspects, and $m$ is the embedding size satisfies $m \ll |\mathcal{V}|$.
\end{definition}

With the learned embeddings, probabilities in Eq.~\ref{eq:form_prob} can be computed in a straightforward manner as described in Sect.~\ref{sect:MHNE}. Therefore, the underlying aspects in the temporal network can be revealed. Note that the learning of multiple aspects and the temporal network embedding are actually complement each other. Especially, we emphasize that the temporal networks are more suitable for multi-aspect embedding learning than the static network for the detailed information in the formation process of temporal networks is essential to induce node aspects. 

\section{Methodology}
\subsection{Hawkes Process}
\label{sect:hawkes_process}
Hawkes process is a typical temporal point process in modeling discrete events considering the temporal decay effect of history events, whose conditional intensity function is defined as follows,
\begin{equation}
	\lambda(t)=\mu(t)+\int_{-\infty}^{t}\mathcal{J}(t-s)\mathrm{d}N(s),
\label{eq: HakesProcess}
\end{equation}
where $\mu(t)$ is the base intensity of a particular event, showing the spontaneous event arrival rate at time $t$, $\mathcal{J}(\cdot)$ is a kernel function that models the time decay effect of history events on the current event, which is usually in the form of an exponential function, and $N(s)$ denotes the number of events until time $s$.

The conditional intensity function of Hawkes process shows that the occurrence of current event does not only depend on the event of last time step, but is also influenced by the historical events with time decay effect. Such property is desirable for modeling the neighborhood formation sequences, for the current neighbor formation can be influenced with higher intensity by the more recent events. 

To model the evolution of an edge based on its source node's various neighbor nodes, multivariate Hawkes process is adopted~\cite{zuo2018embedding}, that is,
\begin{equation}
	\lambda_d(t)=\mu_d(t)+\sum_{d^{\prime}=1}^{D}\int_{-\infty}^{t}\mathcal{J}_{d^{\prime}d}(t-s)\mathrm{d}N_{d^{\prime}}(s),
\label{eq: Multi-variable HawkesProcess}
\end{equation} 
where the conditional intensity function is designed for each event type as one dimension, such as the $d$ or $d^\prime$ dimension. The excitation effects are indeed a sum over all the historical events with different types, captured by an excitation rate $\alpha_{d^\prime, d}$ between dimension $d$ and $d^\prime$, formally, $\mathcal{J}_{d^{\prime}d}(t-s)=\alpha_{d^{\prime}d}\mathcal{J}(t-s)$.

When modeling multi-aspect temporal network, the neighborhood formation sequence can be further decomposed into multiple event sequences according to their underlying aspects. Therefore we resort to the Mixture of Hawkes Process~(MHP), where event sequences driven by various aspects can be modeled simultaneously.

Suppose that there are $K$ different aspects that driven the occurrence of current event~({i.e.} edge), each aspect can be modeled as a component of the MHP. Extended from the Eq.~\ref{eq: Multi-variable HawkesProcess}, we have the intensity function of the current event motivated by aspect $k$ as follows,
\begin{equation}
	\lambda_d^k(t)=\mu_d^k(t)+\sum_{d^{\prime}}^{D}\alpha_{d^{\prime}d}^k\int_{-\infty}^{t}\mathcal{J}(t-s)\mathrm{d}N_{d^{\prime}}^k(s),
	\label{eq: mixture of hawkes}
\end{equation}
where  $\mu_d^k(t)$ denotes the base intensity on current event $d$ excited by the $k$-th aspect, $N_{d}^k(t)$ is the number of events that driven by the $k$-th aspect until time $t$, $\alpha_{d^{\prime}d}^k$ denotes the mutual excitation effect between dimension $d^{\prime}$ and $d$ in terms of the $k$-th aspect. Therefore the intensity function of MHP is $\lambda_d(t)=\pi^k \lambda_d^k(t),$ where $\pi^k$ is the mixture weight.

\subsection{Mixture of Hawkes based temporal Network Embedding}
\label{sect:MHNE}

\begin{figure*}[h!]
	\centering
	\includegraphics[width=0.88\textwidth]{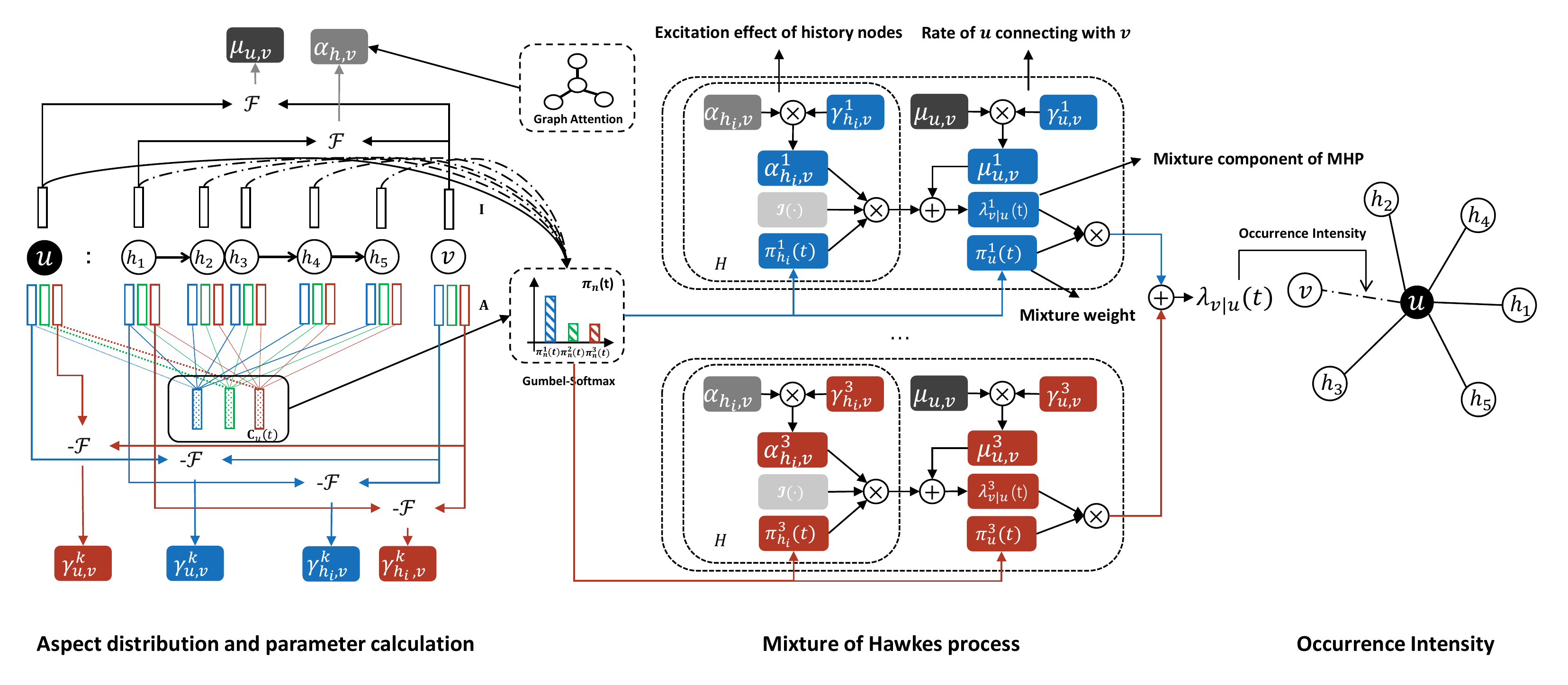}
	\vspace{-0.3cm}
	\caption{Framework of MHNE}
	\vspace{-0.1cm}
	\label{Fig: MHNE Framework}
\end{figure*}
As discussed above, we firstly model the aspect driven neighborhood formation sequences via Mixture of Hawkes Process~(MHP), which is then adapted to learn embeddings. In this way, we can tackle the multi-aspect temporal network embedding problem. Now we formally propose our model named \textbf{M}ixture of \textbf{H}awkes temporal \textbf{N}etwork \textbf{E}mbedding~(MHNE). 

Assume there are $K$ aspects in the multi-aspect temporal network, the intensity function of the MHP for node $v$ connecting to node $u$ at time $t$ can be written as,
\begin{equation}
	\lambda_{v|u}(t) = \sum_{k}\pi_u^k(t)\lambda_{v|u}^k(t),
	\label{eq: lambdasum}
\end{equation}
where $\pi_u^k(t)$ denotes the probability of node $u$ driven by $k$-th aspect at time $t$~({\it i.e.}, the mixture weight of $k$-th aspect), $\lambda_{v|u}^k(t)$ denotes the intensity of node $v$ connects with $u$ at time $t$ driven by $k$-th aspect, which is a intensity function of multivariate Hawkes process. 

Recall the Definition~\ref{def:def_2}, given the aspect distribution $\mathcal{K}_u(t)$ of the source node $u$, the the probability $p(v|u,t,\mathcal{H}_u(t))$ that $u$ connects to $v$ at time $t$ can be written as Eq.~(\ref{eq:form_prob}). By comparing Eq.~(\ref{eq:form_prob}) and Eq.~(\ref{eq: lambdasum}), it is not hard to find that $\pi_u^k(t)$ and $\lambda_{v|u}^k(t)$ are analog to the $\mathcal{K}_u(t)$ and $p(v|u,t,\mathcal{H}_u(t),k)$ in Eq.~(\ref{eq:form_prob})~respectively, which indicates the MHP intensity function defined above is modeling the aspect driven edge formation. Therefore, the MHP has potential to be adapted to tackle the multi-aspect temporal network embedding problem. In the following, we show how to adapt the MHP with multi-aspect embeddings, by firstly describing how to adapt the $\lambda_{v|u}^k(t)$, then presenting how to obtain the aspect distribution $\pi_u^k(t)$.

\noindent\textbf{Adapted Intensity Function.} The intensity $\lambda_{v|u}^k(t)$ can be formulated according to Eq.~(\ref{eq: mixture of hawkes}), that is,
\begin{equation}
	\tilde{\lambda}_{v|u}^k(t) = \mu_{u, v}^k+\sum_{h\in\mathcal{H}_u(t)}\pi_h^k(t)\alpha_{h, v}^k\mathcal{J}(t-t_h),
	\label{eq: lambda k,short}
\end{equation}
where $\mu_{u, v}^k$ denotes the rate of node $u$ connects to $v$ motivated by $k$-th aspect, $h$ denotes the historical nodes connected to $u$ before time $t$, $\alpha_{h,v}^k$ denotes the excitation effect of history node $h$ on target node $v$ and $\pi_h^k(t)$ denotes the activeness of node $h$ on $k$-th aspect. The $\mu_{u, v}^k$ can be further decomposed into the base rate $\mu_{u, v}$ that node $u$ and node $v$ connecting with each other, and the excitation effect $\gamma_{u, v}^k$ from $k$-th aspect. Then, we have $\mu_{u, v}^k=\mu_{u, v}\gamma_{u, v}^k$. Similarly, $\alpha_{h,v}^k$ can be decomposed into $\alpha_{h, v}\gamma_{h, v}^k$. Based on above consideration, Eq.~(\ref{eq: lambda k,short}) is rewritten as,
\begin{equation}
	\tilde{\lambda}_{v|u}^k(t) = \mu_{u,v}\gamma_{u,v}^k+\sum_{h\in\mathcal{H}_u(t)}\pi_h^k(t)\alpha_{h,v}\gamma_{h,v}^k\mathcal{J}(t-t_h),
	\label{eq: lambda k,long}
\end{equation}
where $\mathcal{J}(t-t_h)=\exp(-\delta_u(t-t_h))$ is the kernel function models time decay effect. Since the influence extent of history can be various between different nodes, we introduct a trainable parameter $\delta_u$ for each node.

Intuitively, the probability that two nodes connect with each other are proportion to their similarity. Thus we define a function $\mathcal{F}: \mathbb{R}^d \rightarrow \mathbb{R}$ that maps the embeddings of two corresponding nodes into a similarity score, with which, we are able to adapt the intensity $\lambda_{v|u}^k$ with embeddings. Specifically, we show $\mu_{u, v}$, $\alpha_{h, v}$ can be computed via $\mathcal{F}$ as
\begin{align}
	\mu_{u, v}=\mathcal{F}(\mathbf{I}_u, \mathbf{I}_v), \\
	\alpha_{h, v} = \mathcal{F}(\mathbf{I}_h, \mathbf{I}_v),
	\label{eq: alpha}
\end{align}
where $\mathbf{I}_u$, $\mathbf{I}_v$ and $\mathbf{I}_h$ are identity embeddings of node $u$, node $v$ and node $h$, respectively.
Although there are many choices, we empirically find that negative Euclidean distance works the best. Therefore, $\mathcal{F}(x, y) = -\left\|x-y\right\|^2$, where $x$ and $y$ are node embeddings. 

Similarly, $\gamma_{u, v}^k$ and $\gamma_{h, v}^k$ can be computed via $\mathcal{F}$ as
\begin{align}
	\gamma_{u, v}^k = -\mathcal{F}(\mathbf{A}_u^k, \mathbf{A}_v^k), \\
	\gamma_{h, v}^k = -\mathcal{F}(\mathbf{A}_h^k, \mathbf{A}_v^k),
\end{align}
where $\mathbf{A}_u^k$, $\mathbf{A}_v^k$ and $\mathbf{A}_h^k$ are aspect embeddings. Notably, in order to maintain the monotonicity of $\mu_{u,v}^k$ and $\alpha_{h,v}^k$, a negative sign is added before $\mathcal{F}$.

Finally, the intensity function $\lambda_{v|u}^k(t)$ can be parameterized with the embedding matrices $\{\mathbf{I}, \mathbf{A}\}$. Since intensity function should take positive value when regarded as a rate per unit time, we apply $\exp(\cdot)$ to convert $\tilde{\lambda}_{v|u}^k(t)$ into positive value, that is,
\begin{equation}
	\lambda_{v|u}^k(t) = \exp(\tilde{\lambda}_{v|u}^k(t)).
	\label{eq: exp lambda}
\end{equation}

\noindent\textbf{Graph Attention Mechanism.} With the advent of Graph Convolution Networks, the idea of update node embeddings with its neighbors' information is widely adopted and gained success in many fields ~\cite{li2018deeper}. Velivckovic et al. ~\cite{velivckovic2017graph} proposed a graph attention mechanism of calculating different weights for this information aggregation process, which to some extent denotes the closeness of neighbors and the source. Intuitively, the closer the history node $h$ is to source node $u$, the more convincing that excitation effect should be. For example, if $h_1$ in Fig.~\ref{Fig: Toy Example} is closer to $u$ than $h_2$, then the excitation effect of $h_1$ should be emphasized with a larger weight. Driven by this idea, we adopt the graph attention mechanism to describe the relationship between history nodes and the source node. Follow ~\cite{velivckovic2017graph} we have,

\begin{equation}
	\text{attn}_{u, h} = \frac{\exp(\textbf{LeakyReLU}(\vec{a}^T(\mathbf{W}\mathbf{I}_u|\mathbf{W}\mathbf{I}_h)))}{\sum_{h^{\prime}}\exp(\textbf{LeakyReLU}(\vec{a}^T(\mathbf{W}\mathbf{I}_u|\mathbf{W}\mathbf{I}_{h^{\prime}})))},
\end{equation}
where $\mathbf{W}$ and $\vec{a}$ are trainable parameters. With $\text{attn}_{u, h}$, the $\alpha_{h, v}$ can be computed more precisely, that is,
\begin{equation}
	\alpha_{h, v} = \text{attn}_{u, h} \times \mathcal{F}(\mathbf{I}_h, \mathbf{I}_v).
\end{equation}

\noindent\textbf{Aspect Distribution Calculation.} Here we discuss the problem of calculating the aspect distribution for each node. Commonly, one takes the behavior of one's company, which means if we take the neighborhood sequence of a source node as the context, then we can infer the activeness of aspects for both the source node and the history nodes based on the context. For example, $u$ might be a researcher focusing on {\it Database} and thus its neighbors~({i.e.} co-authors) in a certain time window might mostly come from this field. Therefore, by observing the neighborhood sequence of $u$ in a certain time window, we can infer that $u$ and nodes in its neighborhood sequence are focusing on {\it Database} at that time. We formally define the context embedding $C_u^k(t)$ for node $u$ with neighborhood formation sequence $\mathcal{H}_u(t)$ as follows,
\begin{equation}
	C_u^k(t) = \frac{1}{2}\left(\frac{\sum_{h\in\mathcal{H}_u(t)}\mathcal{J}(t-t_h)\mathbf{A}_h^k}{|\mathcal{H}_u(t)|}+\mathbf{A}_u^k\right).
\end{equation}
Note that the context embeddings of history nodes can be obtained accordingly.

After $K$ context embeddings (one for an aspect) are obtained, we apply Gumbel-Softmax trick~\cite{jang2016categorical} to assign a probability distribution over these aspects. Notably, since we use negative Euclidean distance as similarity metric, the $\log$ operation in conventional Gumbel-Softmax trick is canceled. Taking history node $h \in \mathcal{H}_u(t) \cup \{u\}$ as an example, we have
\begin{equation}
	\pi_h^k(t) = \frac{\exp(\mathcal{F}(\mathbf{I}_h, C_u^k(t))+g_k)/\tau_h)}{\sum_{k^{\prime}}\exp(\mathcal{F}(\mathbf{I}_h, C_u^{k^\prime}(t))+g_{k^{\prime}})/\tau_h)}, 
\end{equation}
where $g_k$ is the gumbel noise sampled from $Gumbel(0,1)$ distribution as,
\begin{align}
	g_k=-\log(-\log(u_k)), \qquad u_k\sim \text{Uniform}(0, 1).
\end{align}
Note that the $\pi_u^k(t)$ can be obtained accordingly.

Although Gumbel-Softmax trick has been used as a workaround to approximate differentiable hard selection over categorical distribution~\cite{park2020unsupervised}, in our work, it is applied for distribution calculation for mainly two reasons. Firstly, although the aspects of a node is highly correlated to its context, but is not determined by it. Therefore, the uncertainty introduced by the gumbel noise is beneficial for us to obtain an aspect distribution rather than determined aspects. Secondly, the temperature parameter $\tau_h$ is a node dependent parameter which controls the extend of approximation between the output of Gumbel-Softmax and a one-hot vector. Since some nodes are mainly focused on an aspect, while some others are not, the node dependent temperature parameter is essential to model the different shapes of the aspect distributions.


\subsection{Model Optimization}
With $\pi_u^k$ and $\lambda_{v|u}^k(t)$, we can compute the $\lambda_{v|u}(t)$ with Eq.~{\ref{eq: lambdasum}}. Then, the probability of target node $v$ connects to source node $u$ at time $t$ can be computed as,
\begin{equation}
	p(v|u, t, \mathcal{H}_u(t)) = \frac{\lambda_{v|u}(t)}{\sum_{v^{\prime}}\lambda_{v^{\prime}|u}(t)}.
	\label{eq: probability softmax}
\end{equation}
Then the log likelihood can be written as follows,
\begin{equation}
	\log \mathcal{L}=\sum_{u\in\mathcal{V}}\sum_{v\in\mathcal{H}_u}\log p(v|u, t, \mathcal{H}_u(t)).
\end{equation}

As the exponential function we introduced in Eq.~(\ref{eq: exp lambda}), Eq.~(\ref{eq: probability softmax}) is indeed a Softmax unit applied to $\tilde\lambda_{v|u}$, which can be optimized approximately via negative sampling~\cite{mikolov2013distributed}. Follow ~\cite{zuo2018embedding}, we sample negative node $v^i$ that shares no links with $v$ according to the degree distribution $p_n(v)\sim d_v^{3/4}$. $d_v$ is the degree for node $v$. Then we give the objective function of MHNE as,
\begin{equation}
	\log \sigma(\tilde{\lambda}_{v|u})+\sum_{i=1}^N\mathbb{E}_{v^i\sim p_n(v)}[-\log \sigma(\tilde{\lambda}_{v^i|u}(t))],
\end{equation}
where $N$ is the number of negative nodes and $\sigma(x)=\exp(x)/(1+\exp(x))$ is the sigmoid function.

We adopt batch gradient descent to optimize the above objective function. In each iteration, we sample a mini-batch of edges with timestamps and fixed length of recently formed neighbors of the source node to update the parameters.

\section{Experimental Setup}
To demonstrate the effectiveness of the proposed MHNE, we conduct extensive experiments to answer the following questions.






RQ1: Can MHNE accurately preserve network structures in embedding space and obtain embeddings with better quality?

RQ2: Can MHNE efficiently capture the temporal pattern of dynamic networks thus precisely predict future events?

RQ3: Is the learned aspect embeddings focusing on certain aspects? Can MHNE provide clues for aspect driven edge formation via MHP?

RQ4: Can MHNE benefit from the incorporated Gumbel-Softmax trick and graph attention mechanism?

RQ5: How do important parameters, including the history length $H$ and number of aspects $K$, influence model performance?

To answer these questions, we choose eight real-world networks and use five baseline methods for comparison.

\subsection{Datasets} 
We examine the effectiveness of our proposed method MHNE on eight publicly available real-world networks with diverse sizes, including a coauthor network (DBLP), two friendship networks (Wosn, Digg), three trust networks (Epinions, BtcAlpha, BtcOtc), a citation network (HepTh) and a network of autonomous systems(Tech). The statistics of these networks are reported in Table~\ref{Tab: Dataset statistics}.
\begin{table}[]
	\caption{Dataset statistics}
	\begin{tabular}{@{}lrrr@{}}
		\toprule
		& \multicolumn{1}{r}{\#nodes} & \multicolumn{1}{r}{\#static edges} & \multicolumn{1}{r}{\#temporal edges} \\ \midrule
		BtcAlpha         & 3748                        & 19139                              & 19139                                \\
		BtcOtc           & 5853                        & 30540                              & 30540                                \\
		HepTh            & 22721                       & 2424641                            & 2651144                              \\
		DBLP             & 27563                       & 129054                             & 203579                               \\
		Techas           & 34761                       & 87719                              & 137966                               \\
		Wosn             & 60663                       & 594540                             & 762268                               \\
		Digg             & 61061                       & 1231934                            & 1403976                              \\
		Epinions         & 119130                      & 813694                             & 813694                               \\\bottomrule
	\end{tabular}
\label{Tab: Dataset statistics}
\end{table}

\subsection{Baselines} 

We compare MHNE with five state-of-the-art~(SOTA) network embedding methods, including static network embedding such as LINE, DeepWalk and asp2vec, as well as temporal network embedding methods such as HTNE and M$^2$DNE. Notably, asp2vec is a SOTA multi-aspect embedding method.

\textbf{LINE}~\cite{tang2015line}: LINE learns embedding by preserving the first-order and second-order proximities for nodes in the network.

\textbf{DeepWalk}~\cite{perozzi2014deepwalk}: DeepWalk first generates random walks from a network, which are treated as sentences and subsequently fed into the Skip-gram~\cite{mikolov2013distributed} model for embedding learning.

\textbf{Asp2vec}~\cite{park2020unsupervised}: This method extends DeepWalk by learning multiple embeddings for a single node in a differentiable way.
 
\textbf{HTNE}~\cite{zuo2018embedding}: HTNE models the neighborhood formation sequence of temporal networks to learn node embeddings. 

\textbf{M$^2$DNE}~\cite{lu2019temporal}: M$^2$DNE extends HTNE by taking both micro and macro dynamics into account.


\subsection{Parameter settings}
In terms of LINE, we employ both first-order and second-order proximities to learn node embeddings, for we empirically observe better results than using just first-order or second-order proximity. The number of walks per node, walk length and window size are set to 10, 80, 3 for DeepWalk and asp2vec. Specifically, the number of aspects is set to 4 for both asp2vec and our method. The history length is set to 5 for our method, HTNE and M$^2$DNE. 

We run each model with varying dimensions of learned embeddings. Specifically, we set the embedding dimension $dim$ as $100$, $200$ and $500$ respectively. Notably, the final embedding is obtained by concatenating identity embedding along with all aspect embeddings for multi-aspect embedding methods such as asp2vec and our method. For a fair comparison, the size of identity/aspect embedding is set to $dim/(K+1)$, where $K$ is the number of aspects. Moreover, as we obtain two embeddings for LINE by employing first and second order proximity, their dimensions are both set to $dim/2$ to ensure the concatenated vector has the same size as others.
\section{Experimental Results}
\subsection{Embedding Quality~(RQ1)}
\begin{table*}[h]
\caption{Link Prediction Results}
\vspace{-0.2cm}
\resizebox{0.95\linewidth}{!}{
\begin{tabular}{l|c|cccccc|cccccc}
\toprule
\multirow{2}{*}{Dataset} & \multirow{2}{*}{dim} & \multicolumn{6}{c|}{f1-macro}                                                                & \multicolumn{6}{c}{AUC-ROC}                                                                \\ \cline{3-14} 
                         &                      & LINE            & asp2vec      & DeepWalk        & HTNE   & M$^2$DNE          & MHNE            & LINE         & asp2vec      & DeepWalk        & HTNE   & M$^2$DNE           & MHNE            \\ \midrule
DBLP                     & \multirow{8}{*}{100} & 0.7751          & 0.9389       & 0.9073          & 0.9335 & {\ul 0.9443}   & \textbf{0.9594} & 0.8459       & 0.9799       & 0.9666          & 0.9807 & {\ul 0.9851}    & \textbf{0.9886} \\
Wosn                     &                      & 0.8168          & 0.8991       & 0.8896          & 0.9005 & 0.9136         & \textbf{0.9323} & 0.8889       & 0.9593       & 0.9526          & 0.9620  & {\ul 0.9691}    & \textbf{0.9774} \\
Digg                     &                      & 0.8034          & 0.8647       & {\ul 0.8862}    & 0.8444 & 0.8701         & \textbf{0.8871} & 0.8710        & 0.9348       & {\ul 0.9507}    & 0.9026 & 0.9404          & \textbf{0.9522} \\
Epinions                 &                      & 0.7941          & 0.9114       & 0.9228          & 0.8916 & {\ul 0.9239}   & \textbf{0.9266} & 0.8590        & 0.9649       & 0.9686          & 0.9580  & {\ul 0.9745}    & \textbf{0.9749} \\
Tech                     &                      & \textbf{0.9172} & 0.8882       & 0.7919          & 0.8919 & 0.9043         & {\ul 0.9086}    & \textbf{0.9685}& 0.9511       & 0.8709          & 0.9534 & 0.9622          & {\ul 0.9630}     \\
HepTh                    &                      & 0.8602          & 0.7981       & \textbf{0.9277} & 0.8556 & 0.8946         & {\ul 0.9222}    & 0.9299       & 0.8776       & \textbf{0.9792} & 0.9307 & 0.9569          & {\ul 0.9737}    \\
BtcAlpha                 &                      & 0.8127          & 0.8919       & 0.8801          & 0.8834 & {\ul 0.9093}   & \textbf{0.9210}  & 0.8838       & 0.9514       & 0.9420           & 0.9453 & {\ul 0.9646}    & \textbf{0.9712} \\
BtcOtc                   &                      & 0.8194          & 0.8954       & 0.8782          & 0.8859 & {\ul 0.9177}   & \textbf{0.9285} & 0.8897       & 0.9557       & 0.9380           & 0.9465 & {\ul 0.9709}    & \textbf{0.9731} \\ \hline
DBLP                     & \multirow{8}{*}{200} & 0.8229          & {\ul 0.949}  & 0.9316          & 0.9298 & 0.9426         & \textbf{0.9627} & 0.8983       & 0.9830        & 0.9765          & 0.9791 & {\ul 0.984}     & \textbf{0.9898} \\
Wosn                     &                      & 0.8204          & {\ul 0.9147} & 0.8840           & 0.9016 & 0.9116         & \textbf{0.9311} & 0.8946       & {\ul 0.9690}  & 0.9459          & 0.9623 & 0.9685          & \textbf{0.9770}  \\
Digg                     &                      & 0.8428          & 0.8737       & {\ul 0.8808}    & 0.8420  & 0.8707         & \textbf{0.8852} & 0.9076       & 0.9409       & {\ul 0.9469}    & 0.9184 & 0.9404          & \textbf{0.9516} \\
Epinions                 &                      & 0.8760           & 0.9123       & 0.9199          & 0.8882 & {\ul 0.9209}   & \textbf{0.9223} & 0.9346       & 0.9645       & 0.9676          & 0.9534 & {\ul 0.9724}    & \textbf{0.9730}  \\
Tech                     &                      & {\ul 0.9105}    & 0.8917       & 0.7994          & 0.8865 & 0.9018         & \textbf{0.9137} & {\ul 0.9627} & 0.9527       & 0.8766          & 0.9504 & 0.9614          & \textbf{0.9670}  \\
HepTh                    &                      & 0.8413          & 0.8563       & {\ul 0.9174}    & 0.8529 & 0.8954         & \textbf{0.9224} & 0.9169       & 0.9278       & {\ul 0.9735}    & 0.9299 & 0.9582          & \textbf{0.9737} \\
BtcAlpha                 &                      & 0.8492          & {\ul 0.9092} & 0.8792          & 0.8763 & 0.9069         & \textbf{0.9256} & 0.9244       & {\ul 0.9647} & 0.9394          & 0.9414 & 0.9614          & \textbf{0.9734} \\
BtcOtc                   &                      & 0.8507          & 0.9112       & 0.8790           & 0.8756 & {\ul 0.9134}   & \textbf{0.9321} & 0.9231       & {\ul 0.9689} & 0.9385          & 0.9400   & 0.9681          & \textbf{0.9743} \\ \hline
DBLP                     & \multirow{8}{*}{500} & 0.8228          & 0.9406       & {\ul 0.9423}    & 0.9250  & 0.9383         & \textbf{0.9611} & 0.8986       & 0.9782       & 0.9798          & 0.9765 & {\ul 0.9823}    & \textbf{0.9892} \\
Wosn                     &                      & 0.8037          & 0.9045       & 0.8636          & 0.8974 & {\ul 0.9115}   & \textbf{0.9215} & 0.8784       & 0.9607       & 0.9305          & 0.9599 & {\ul 0.9690}     & \textbf{0.9728} \\
Digg                     &                      & 0.8507          & {\ul 0.8778} & 0.8690           & 0.8325 & 0.8636         & \textbf{0.8795} & 0.9129       & {\ul 0.9445} & 0.9379          & 0.9110  & 0.9347          & \textbf{0.9465} \\
Epinions                 &                      & 0.8800            & 0.8894       & {\ul 0.9140}     & 0.8802 & \textbf{0.9180} & 0.9116          & 0.9377       & 0.9505       & 0.9632          & 0.9499 & \textbf{0.9714} & {\ul 0.9677}    \\
Tech                     &                      & {\ul 0.9085}    & 0.9030        & 0.8141          & 0.8810  & 0.8969         & \textbf{0.9115} & {\ul 0.9627} & 0.9615       & 0.8900            & 0.9474 & 0.9595          & \textbf{0.9671} \\
HepTh                    &                      & 0.8479          & 0.8650        & {\ul 0.9040}     & 0.8530  & 0.8965         & \textbf{0.9156} & 0.9251       & 0.9334       & {\ul 0.9652}    & 0.9283 & 0.9585          & \textbf{0.9710}  \\
BtcAlpha                 &                      & 0.8243          & {\ul 0.9041} & 0.8758          & 0.8644 & 0.8996         & \textbf{0.9249} & 0.9032       & {\ul 0.9610}  & 0.9392          & 0.9326 & 0.9576          & \textbf{0.9725} \\
BtcOtc                   &                      & 0.8123          & {\ul 0.9190}  & 0.8751          & 0.8625 & 0.9052         & \textbf{0.9289} & 0.8916       & {\ul 0.9714} & 0.9399          & 0.9300   & 0.9593          & \textbf{0.9741} \\ \bottomrule
\end{tabular}}
\label{Tab: Link prediction}
\end{table*}

Link prediction is suggested to be the primary choice of evaluating the unsupervised network embedding methods~\cite{abu2017learning, epasto2019single}. Other downstream tasks such as node classification might involve low quality labels that cannot represent the real status of nodes, which can lead to unreliable results. Moreover, as we assume that each node has multiple aspects, it is difficult to categorize each node with a single label. Therefore in order to validate the effectiveness of MHNE in accurately preserving the network structure in the embedding space, we compared the link prediction results based on the embeddings learned from the embedding methods. 

Specifically, we aim to determine whether there is an edge between two nodes based on its representation, which is constructed by calculating the element-wise absolute difference between its two corresponding nodes' embeddings. In our experiment, we first randomly mask $20,000$ static edges from the networks, except for BtcAlpha and BtcOtc, where we randomly mask $5,000$ static edges. Those masked edges are taken as positive instances. Meanwhile, we randomly sample equal number of false edges that do not exist in the network as negative instances. Then, a \textit{Logistic Regression} classifier is trained on $50\%$ of the mixed positive and negative instances, and we report the {\it macro-f1} and {\it AUC-ROC} based on the remaining instances in Table~\ref{Tab: Link prediction}.

From the results in Table~\ref{Tab: Link prediction}, we can easily find that MHNE outperforms the baseline methods in nearly all the cases in terms of both metrics. This promising results suggest that we can better recover the formation process of networks by modeling the multi-aspects of nodes in a dynamic setting, and can obtain node embeddings with better quality. We notice that although MHNE outperforms all the baseline methods when the embedding size is set to be 200, there still exist some exceptions. LINE and DeepWalk performs the best on Tech and Hepth respectively when $dim=100$, and M$^2$DNE performs the best on Epinions when $dim=500$. However, MHNE is outperformed by no more than $0.006$ in terms of both f1-macro and AUC-ROC among all these exceptions, showing its robustness.

It is argued that multi-aspect embedding methods are more preferable in modeling social networks~\cite{park2020unsupervised}, since nodes in such networks inherently show multiple aspects. We find that MHNE consistently outperforms all the baseline methods on the two social networks, which is not only in line with the previous findings but also confirms that the temporal information can better reveal the multiple aspects of nodes. 

\subsection{Temporal Node Recommendation~(RQ2)}

\begin{figure}[t!]
	\centering
	\begin{subfigure}[b]{0.235\textwidth}
		\includegraphics[width=\textwidth]{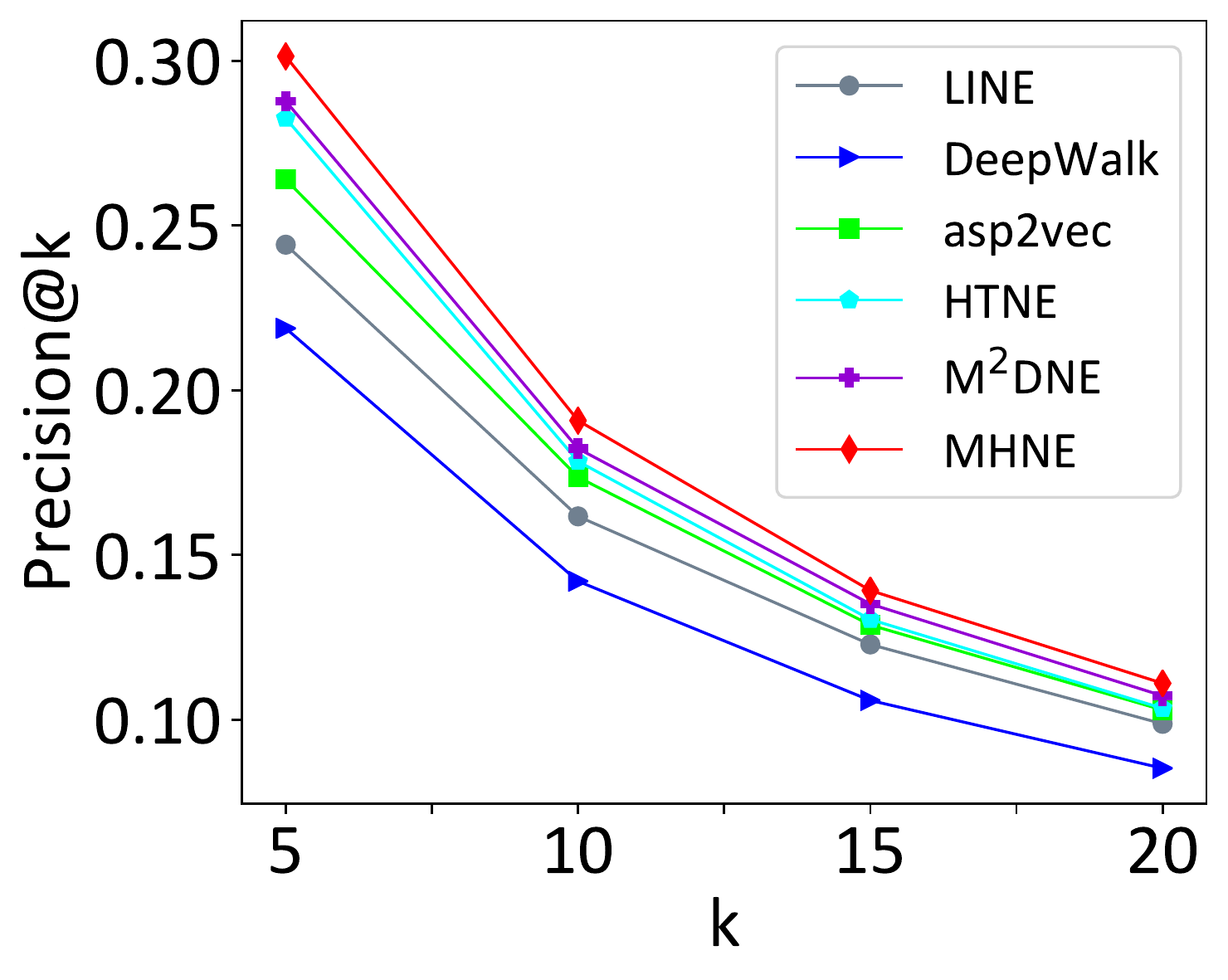}
		\caption{precision@k}
		\label{Fig: TNR precision}
	\end{subfigure}
	\begin{subfigure}[b]{0.235\textwidth}
		\includegraphics[width=\textwidth]{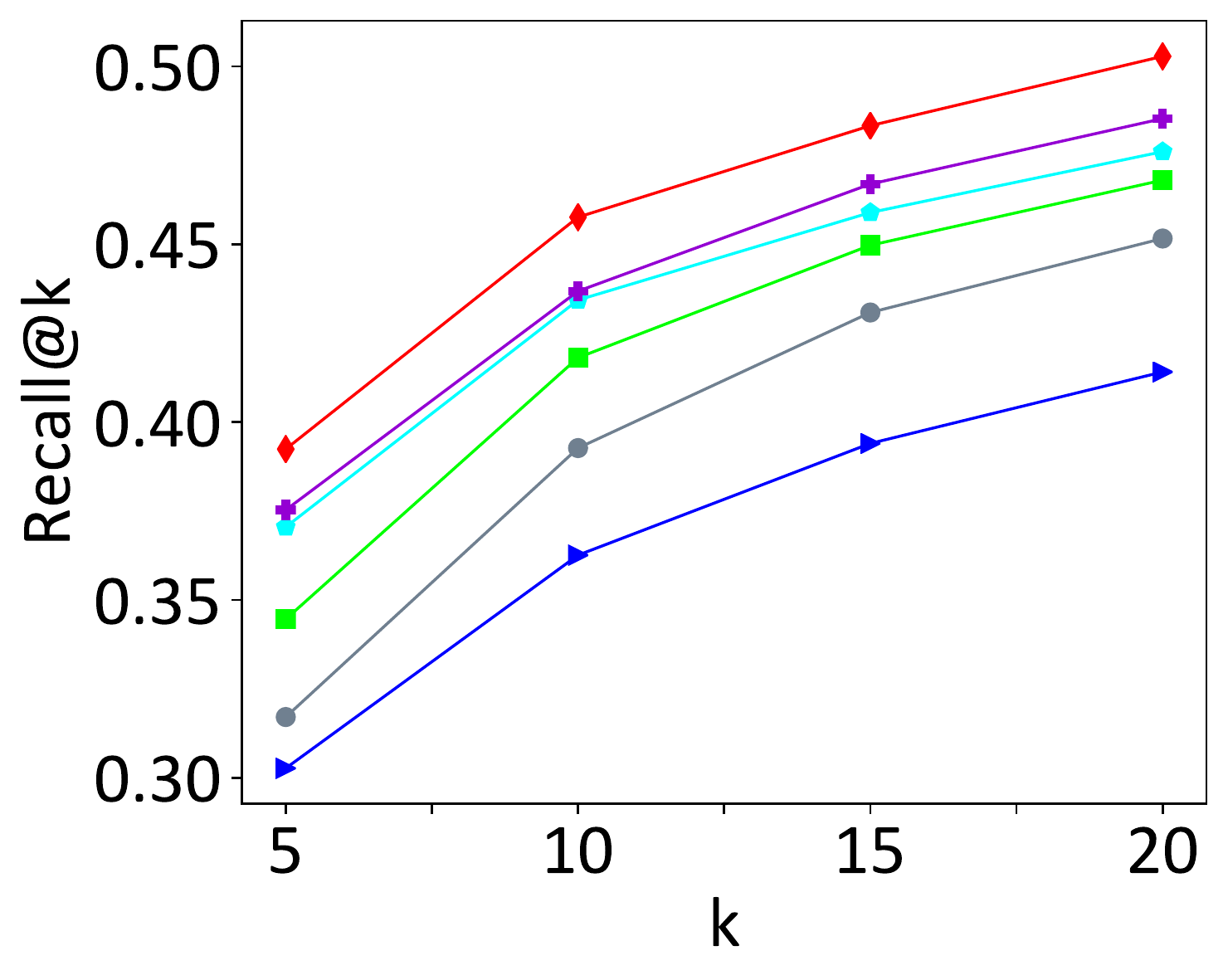}
		\caption{recall@k}
		\label{Fig: TNR recall}
	\end{subfigure}
	\vspace{-0.4cm}
	\caption{Temporal Node Recommendation Results}
	\vspace{-0.4cm}
	\label{Fig: TNR}
\end{figure}

To further evaluate whether our method can better capture the evolution patterns of temporal networks, we conduct the temporal node recommendation experiment on DBLP, \emph{i.e.}, to recommend possible coauthors for a given researcher at a specific time. Specifically, we learn the node embeddings on the network extracted between the time interval $[t^{\prime}, t)$, and use the coauthor relationships established at and after the time $t$ as test set. After the embeddings are learned, we calculate the intensity of building connections with other nodes as the ranking scores for recommendation. Follow the settings in~\cite{lu2019temporal, zuo2018embedding}, we apply inner product as ranking score for DeepWalk and LINE, and apply negative Euclidean distance for asp2vec, HTNE, M$^2$DNE and our method. Afterwards, we derive the top-$k$ candidate nodes at time $t$, and report the results in terms of {\it Precision@k} and {\it Recall@k} in Fig.~\ref{Fig: TNR}.

We can find from Fig.~\ref{Fig: TNR} that our method consistently outperforms the other methods in terms of both metrics. Interestingly, we find the three methods (MHNE, M$^2$DNE, HTNE) which model the formation process outperform the other static embedding methods. This demonstrates that by modeling the dynamic neighborhood formation process, the future events can be predicted more accurately. Moreover, though asp2vec cannot compete with the temporal network embedding methods, it achieve the best performances among all the static embedding methods, which again provides evidence for the necessity in considering the multiple aspects. Therefore, it also emphasizes that multi-aspect embedding and temporal network embedding are two tasks complement each other.

\label{sec: temporal node recommendation}
\subsection{Aspect Embedding Analysis~(RQ3)}
In this experiment, we aim to explore whether our method can capture different aspects of nodes from the two perspectives. \textbf{1)} What are the quality of the learned aspect embeddings? \textbf{2)} Can MHNE capture the diverse aspect driven neighborhood formations?
\subsubsection{Aspect Embedding Quality}
\begin{figure}[t!]
	\centering
	\begin{subfigure}[b]{0.23\textwidth}
		\includegraphics[width=\textwidth]{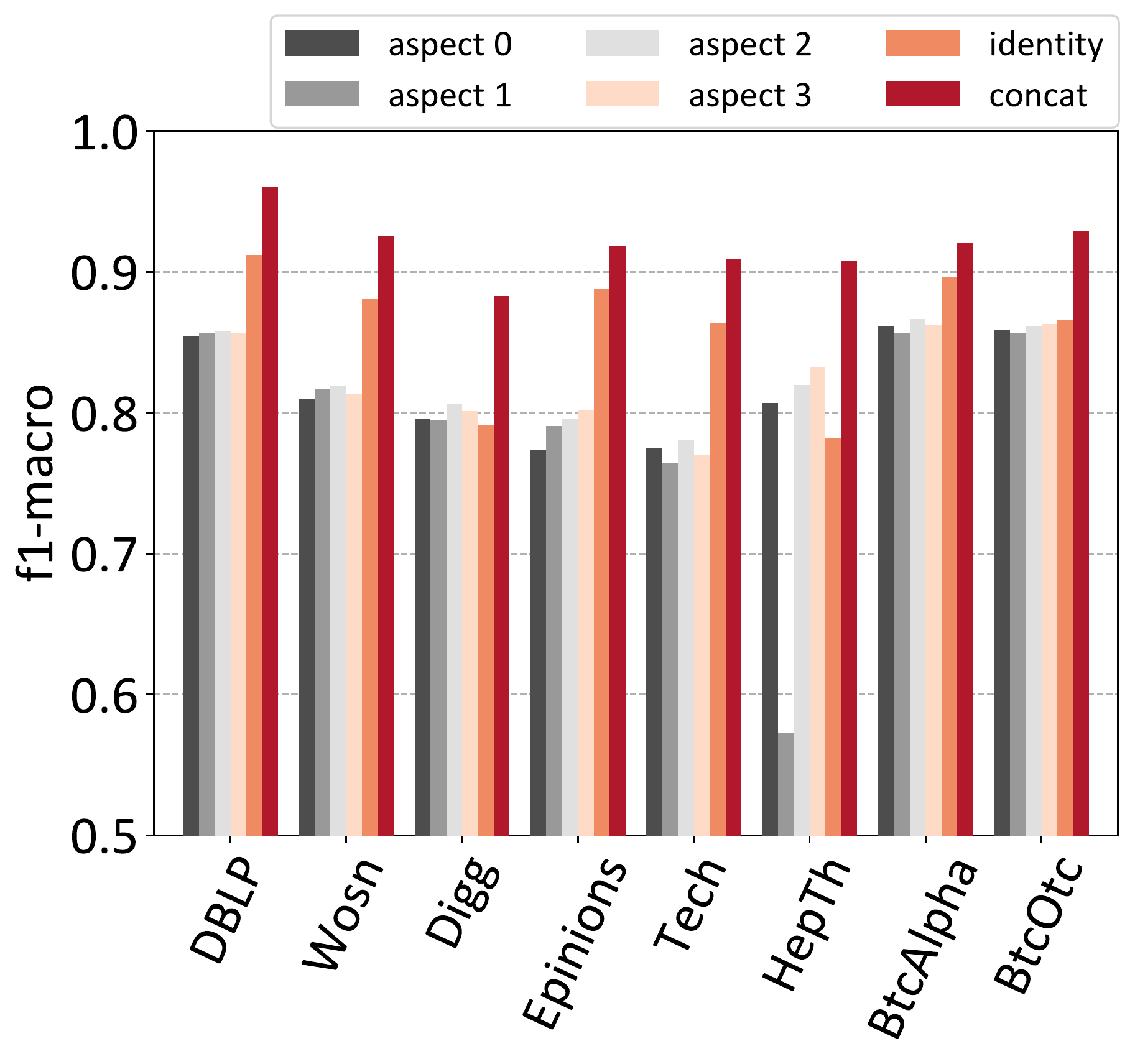}
		\caption{MHNE}
		\label{Fig: Asp Pred MHNE}
	\end{subfigure}
	\begin{subfigure}[b]{0.23\textwidth}
		\includegraphics[width=\textwidth]{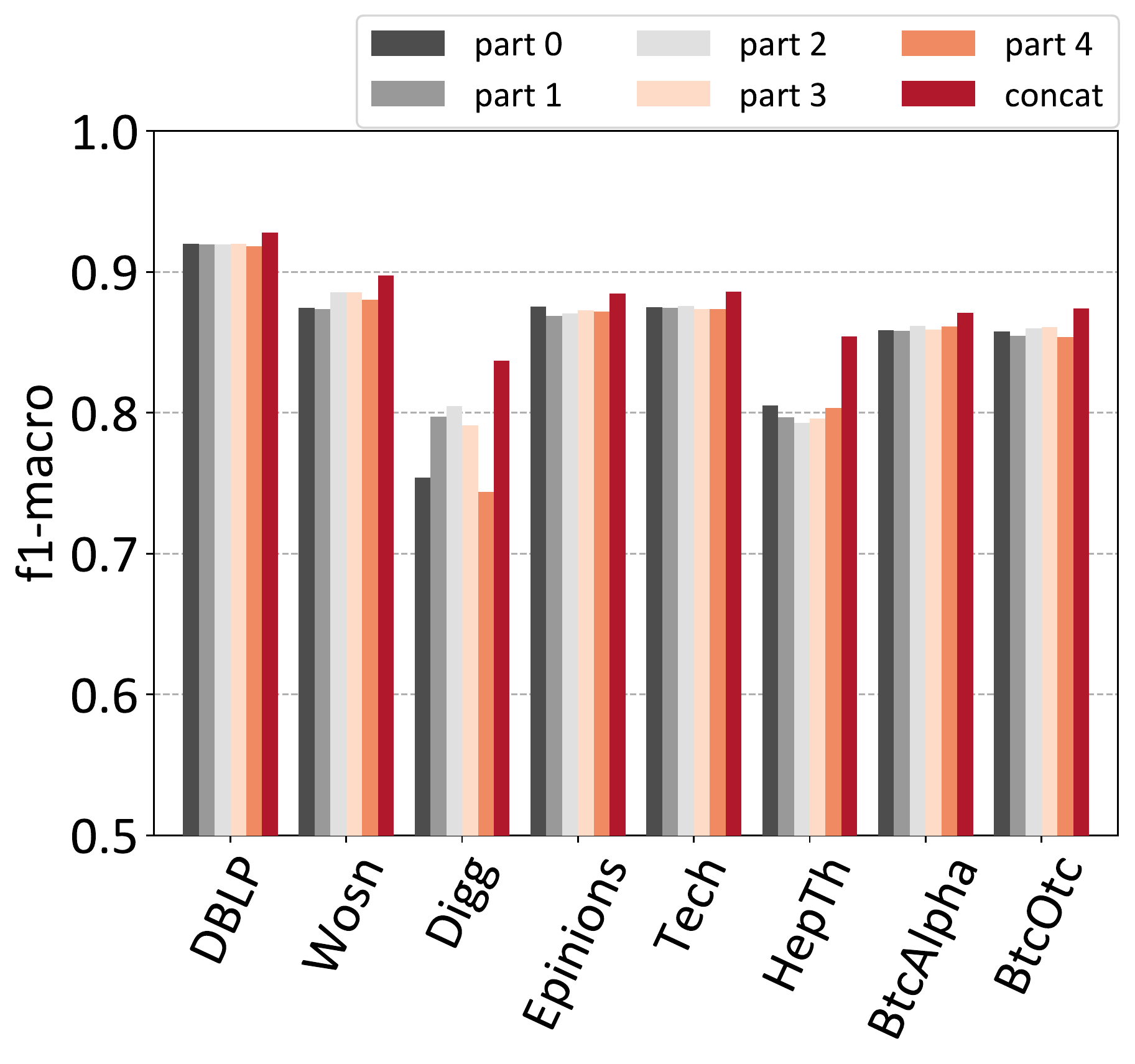}
		\caption{HTNE}
		\label{Fig: Asp Pred HTNE}
	\end{subfigure}
	\vspace{-0.4cm}
	\caption{Link Prediction with Aspect embeddings}
	\vspace{-0.4cm}
	\label{Fig: Aspect Link Prediction}
\end{figure}

To evaluate the quality of the aspect embeddings, we perform the link prediction using only embeddings from certain aspect or identity embeddings and report the macro-f1 in Fig.~\ref{Fig: Aspect Link Prediction}. For comparison, we conduct this experiment on HTNE by splitting its node embeddings into multiple parts that have equal size with the aspect embeddings.

Fig.~\ref{Fig: Asp Pred MHNE} shows that aspect embeddings are significantly outperformed by the concatenated vectors consisting of both aspect embeddings and identity embeddings. Conversely, the performance gap between the embeddings learned from HTNE and their splitted parts is much smaller as illustrated in Figure.~\ref{Fig: Asp Pred HTNE}. Such contrastive results may be due to the fact that the aspect embeddings obtained by MHNE are one-sided and cannot adequately describe the nature of nodes from a macro perspective, since we force each aspect embedding to focus on a certain aspect. Therefore, the performance improvement of concatenating the aspect and identity embeddings is dramatically larger, which demonstrates that we can obtain a more comprehensive description of nodes from its various aspects. 
\subsubsection{Analysis of Aspect-driven Temporal Events}
\begin{figure}[t!]
	\centering
	\begin{subfigure}[b]{0.44\textwidth}
		\includegraphics[width=\textwidth]{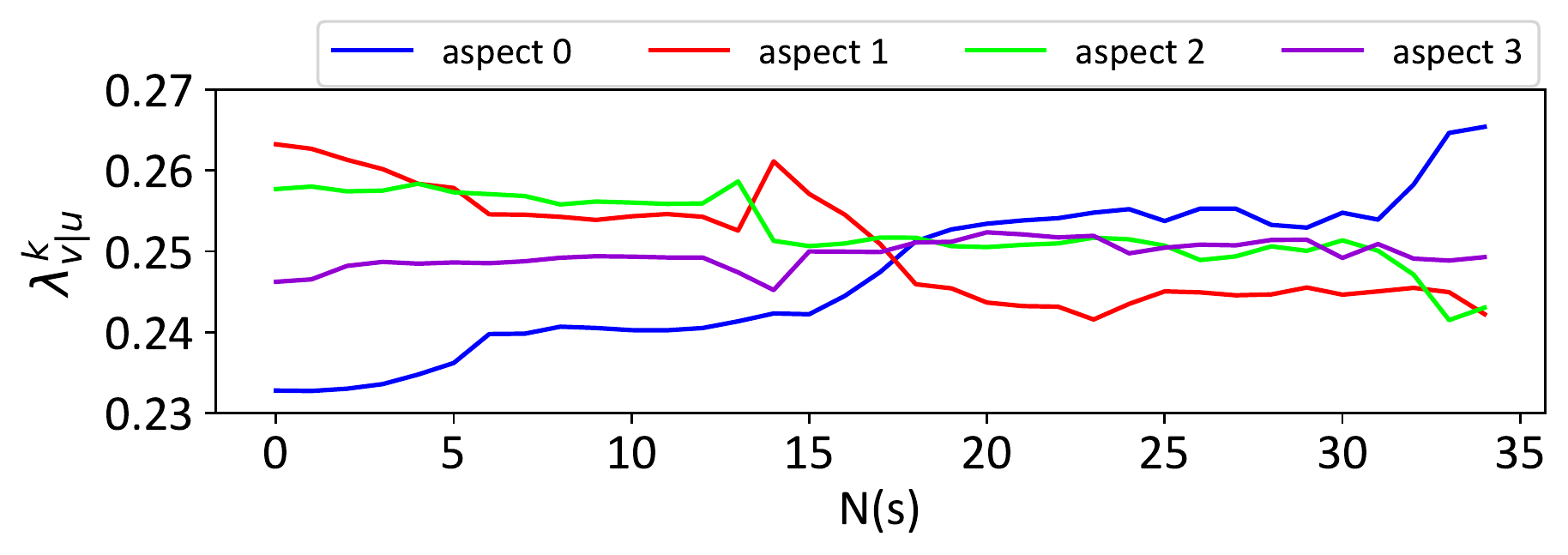}
		\caption{Bolin Ding}
		\label{Fig:case study bolin}
	\end{subfigure}

	\begin{subfigure}[b]{0.44\textwidth}
		\includegraphics[width=\textwidth]{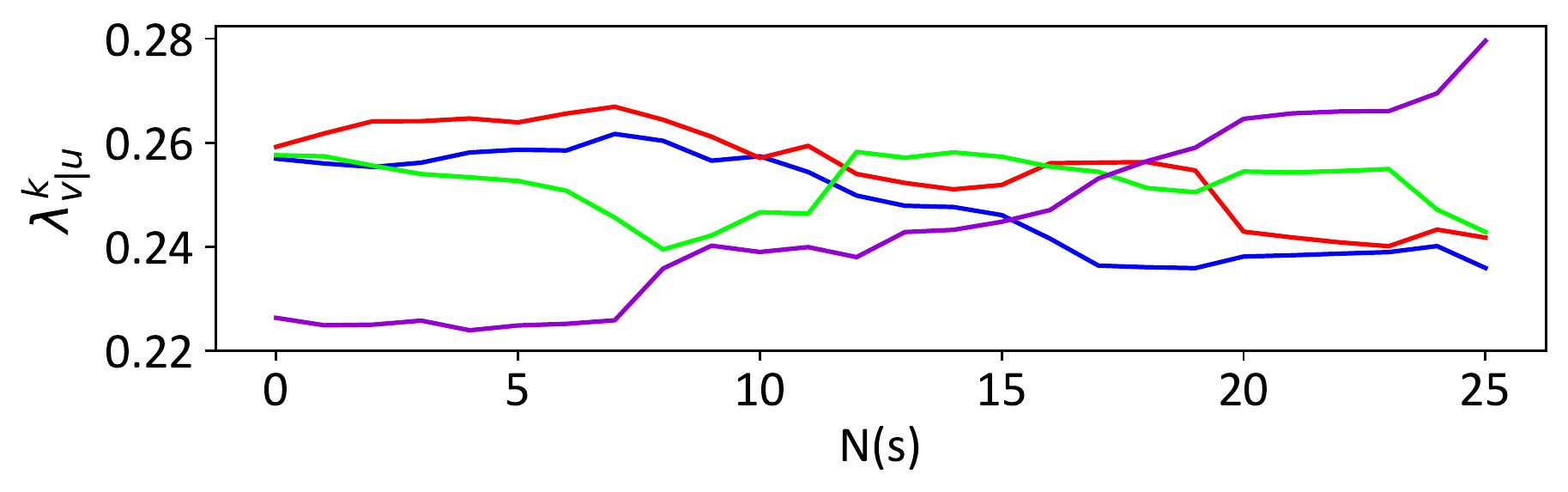}
		\caption{Christopher R\'e}
		\label{Fig:case study christopher}
	\end{subfigure}
	\caption{Aspect-driven Temporal Events}
	\vspace{-0.4cm}
	\label{Fig: Lambda Sequence}
\end{figure}
In order to gain insights on how the learned multiple different aspects affect the occurrence of temporal events, we visualize the intensity excited by aspect $k$, \emph{i.e.}, $\lambda^k_{v|u}$ of the neighborhood sequence. For illustrative purpose, we choose two scholars, \emph{Bolin Ding} and {\it Christopher R\'e} from DBLP who were previously focusing on the {\it Database} field and then shift the research interests to {\it Data Mining} and {\it Artificial Intelligence} respectively. This means that {\it Bolin Ding} and {\it Christopher R\'e} might collaborate with other researchers focusing on {\it Database} in the earlier years, while their academic partnership may change accordingly over time.


As illustrated in Fig.~\ref{Fig: Lambda Sequence}, the red lines in both subfigures representing the intensity from aspect $\#1$ has a clear downward trend, while the blue line in Fig.~\ref{Fig:case study bolin} and purple line in Fig.~\ref{Fig:case study christopher} keep rising over time. This indicates that the temporal events of {\it Bolin Ding} and {\it Christopher R\'e} are mostly driven by aspect $\#1$ at the early stages. After that, the occurrence of temporal events are primarily inspired by aspect $\#0$ and aspect $\#3$ for them. Compared with previous analysis, we infer that aspect $\#1$ mainly represents {\it Database} for these two authors, and aspect $\#0$ and aspect $\#3$ might represent {\it Data Mining} and {\it Artificial Intelligence} respectively. These two cases demonstrate that MHNE can capture different driving aspects for the temporal events and can be reflected from the dynamic node status.



\subsection{Ablation Study~(RQ4)}

\begin{figure}[t!]
	\centering
	\includegraphics[width=0.46\textwidth]{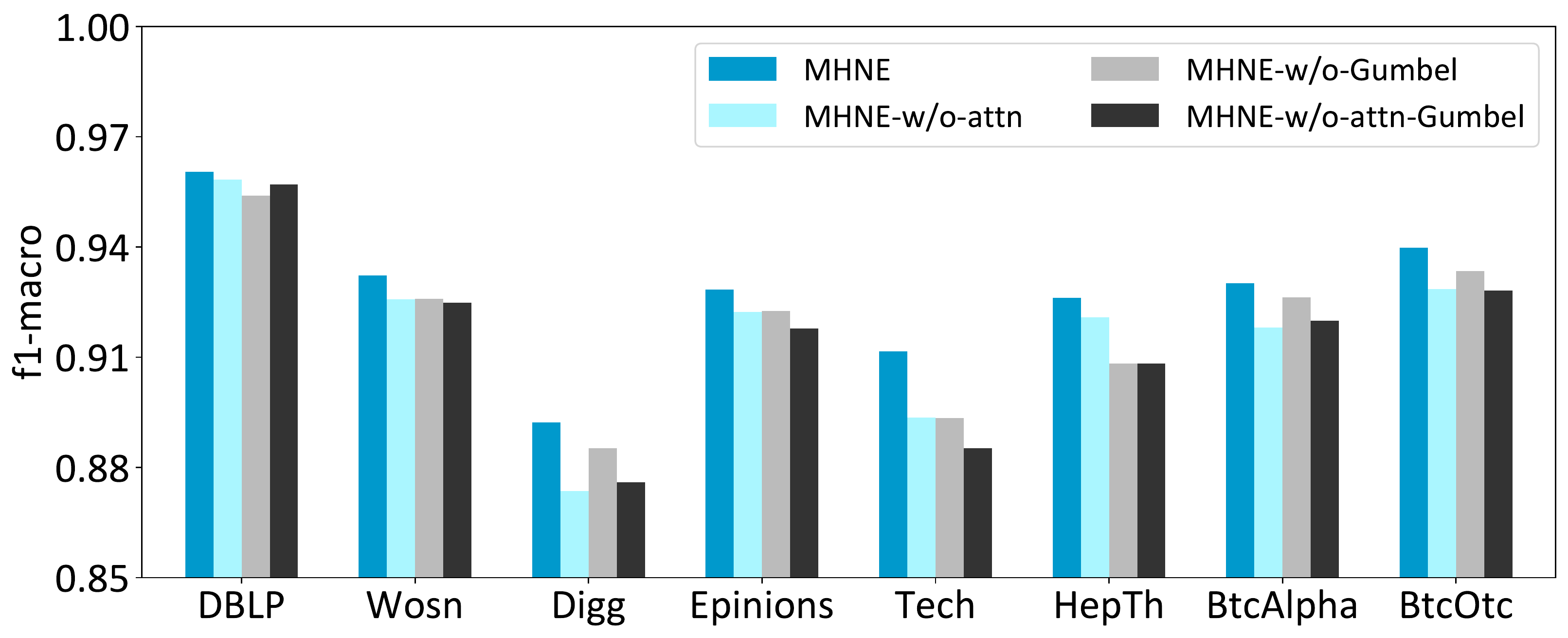}
	\caption{Ablation Study}
	\vspace{-0.2cm}
	\label{Fig: Ablation Study}
\end{figure}

To validate the effectiveness of incorporating graph attention mechanism and Gumbel-Softmax into our model, we construct several submodels that deliberately removes one component or both and test the performances of these model variants. Specifically, we conduct link prediction with these constructed models and report results in terms of the macro-f1 in Fig.~\ref{Fig: Ablation Study}.

From the results we can find MHNE outperforms all the submodels in all the datasets, which demonstrate the rationality of incorporating both graph attention mechanism and Gumbel-Softmax. Among the submodels, MHNE-w/o-attn performs the best on HepTh and MHNE-w/o-Gumbel performs the best on Digg, BtcAlpha and BtcOtc. In the other datasets, MHNE-w/o-attn and MHNE-w/o-Gumbel performs on par with each other. These results verify that MHNE benefits from graph attention mechanism by better capturing the relationship between the source node and history nodes. Moreover, Gumbel-Softmax trick enables MHNE to assign different forms of aspect distributions for different nodes, which also improves model performance. 
In addition, MHNE-w/o-attn-Gumbel performs the worst in most cases.

\subsection{Parameter Sensitivity~(RQ5)} 

For point process based embedding models~\cite{zuo2018embedding, lu2019temporal} and multi-aspect embedding methods~\cite{park2020unsupervised, liu2019single, epasto2019single}, the most important parameters are the history length $H$ and the number of aspects $K$ respectively. In this section, we study the sensitivity of our model by conducting link prediction experiments under different setting of the two parameters, and we report the macro-f1 in Table~\ref{Tab: History Length} and Table~\ref{Tab: Number of Aspects}. Notably, we fix the dimension of concatenated vectors to $200$, thus the size of each aspect is set to $200/(K+1)$.

As shown in Table~\ref{Tab: History Length}, the optimal history length varies in different datasets. While the performance has never be the best when $H=1$, we argue that MHNE benefits for a comparatively longer history length which incorporates more information from the neighborhood sequence. As illustrated in Table~\ref{Tab: Number of Aspects}, we can see that the macro-f1 of MHNE increases along with $K$ and remains stable when $K>=2$. Specifically, we preserve only one aspect embedding for each node when $K=1$, which means that the multi-aspect nature is ignored. Interestingly, MHNE performs the best when $K$ is set to be the largest on the two social networks, which is in accordance with previous discussion that the nodes in social networks inherently exhibits multiple aspects than other networks. This result again supports that our model can benefit from learning multiple aspect embeddings while is robust on different number of aspects across all the datasets.

\begin{table}[t!]
\caption{Link prediction over different history length}
\small
\begin{tabular}{@{}l|ccccc@{}}
\toprule
$H$             & 1      & 2               & 3               & 4               & 5               \\ \midrule
DBLP           & 0.9613 & \textbf{0.9613} & 0.9603          & 0.9596          & 0.9599          \\
Wosn           & 0.9276 & 0.9289          & \textbf{0.9300}   & 0.9290           & 0.9286          \\
Digg           & 0.8885 & 0.8872          & 0.8878          & 0.8882          & \textbf{0.8892} \\
Epinions       & 0.9228 & 0.9243          & 0.9247          & 0.9261          & \textbf{0.9262} \\
Tech           & 0.9119 & 0.9128          & \textbf{0.9143} & 0.9129          & 0.9119          \\
HepTh          & 0.9273 & 0.9287          & \textbf{0.9294} & 0.9288          & 0.9232          \\
BtcAlpha       & 0.9283 & 0.9302          & \textbf{0.9310}  & 0.9293          & 0.9296          \\
BtcOtc         & 0.9378 & 0.9367          & 0.9349          & \textbf{0.9393} & 0.9346          \\ \bottomrule
\end{tabular}
\label{Tab: History Length}
\end{table}

\begin{table}[t!]
\small
\caption{Link prediction over different number of aspects}
\begin{tabular}{@{}l|ccccc@{}}
\toprule
$K$             & 1      & 3      & 4               & 7               & 9               \\ \midrule
DBLP           & 0.9579 & 0.9622 & \textbf{0.9624} & 0.9615          & 0.9601          \\
Wosn           & 0.9186 & 0.9283 & 0.9295          & 0.9312          & \textbf{0.9315} \\
Digg           & 0.8608 & 0.8851 & 0.8825          & 0.8912          & \textbf{0.8923} \\
Epinions       & 0.8966 & 0.9165 & 0.9194          & \textbf{0.9252} & 0.9249          \\
Tech           & 0.9024 & 0.9089 & \textbf{0.9099} & 0.9080           & 0.9086          \\
HepTh          & 0.8757 & 0.9177 & 0.9230           & 0.9240           & \textbf{0.9299} \\
BtcAlpha       & 0.9106 & 0.9260  & 0.9230           & \textbf{0.9262} & 0.9234          \\
BtcOtc         & 0.9184 & 0.9294 & 0.9290           & 0.9292          & \textbf{0.9324} \\ \bottomrule
\end{tabular}
\label{Tab: Number of Aspects}
\end{table}

\section{Related Work}
Network embedding methods represent each node in the network via low-dimensional vectors while preserving network structure, have gained wide attention in recent years~\cite{cui2018survey}, and our work is related to the following categories of studies.

\textbf{Network embedding.} Mikolov et al.~\cite{mikolov2013distributed, mikolov2013efficient} proposes an efficient embedding learning method for words and phrases in natural language. Inspired by their work, several random walk based network embedding methods are developed by making analogy between nodes and words in natural language. For example, DeepWalk~\cite{perozzi2014deepwalk} and node2vec~\cite{grover2016node2vec} use truncated random walks generated from a network as the input of skip-gram model to learn node embeddings. Besides, LINE~\cite{tang2015line} optimizes node embeddings by preserving first-order and second-order proximities in both vertex and embedding space. Graph Convolutional Networks(GCNs) are another approach on representation learning based on the message-passing-receiving mechanism, which is to update a node's representation by aggregating information from its neighbors~\cite{velivckovic2017graph, hamilton2017inductive, kipf2016semi}. However, most GCNs are supervised or semi-supervised method thus has a strong requirement for a large amount of labeled data~\cite{li2018deeper}.

\textbf{Temporal network embedding.} These aforementioned methods all focus on static networks, however, in reality, the majority of networks evolve over time. ~\cite{pareja2020evolvegcn, deng2019learning} adapt conventional GCNs to model temporal networks by modeling multiple snapshots derived from different time windows. ~\cite{goyal2018dyngem} employs deep autoencoders on derived network snapshots, and the structure of the autoencoder evolves along with the growing of network scale. In contrast to the snapshot based methods, some efforts~\cite{zuo2018embedding, lu2019temporal} view the formation of temporal networks as adding nodes and edges and they trace back the formation process by tracking the neighborhood formation of each node. For example, zuo et al.~\cite{zuo2018embedding} proposes HTNE by modeling the neighborhood formation sequence with a multivariate Hawkes process. M$^2$DNE ~\cite{lu2019temporal} extends HTNE by incorporating both micro and macro dynamics of the network, namely neighborhood formation and network scale. Mei et al.~\cite{mei2017neural} proposed Neural Hakes process by using a continuous LSTM model to better reveal the sophisticated mutual effect between history and future events. However, the underlying mechanism of how temporal edges are formed driven by different aspect is ignored in prior work, which is crucial to model the multi-aspect temporal networks.

\textbf{Multi-aspect embedding}. There exists some recent research~\cite{park2020unsupervised, liu2019single, epasto2019single, wang2019mcne} learning multiple embedding vectors for each node in order to capture the multifaceted nature of nodes. Liu et al.~\cite{liu2019single} proposes PolyDW by extending DeepWalk to multi-aspect embedding setting, it determines the aspect distribution for each node via matrix factorization before embedding learning. Then the activated aspect of both target node and context node is sampled from the computed distribution, respectively. Splitter~\cite{epasto2019single} firstly creates a persona graph from the original network via a cluster algorithm that maps each node to one or multiple personas, on which a random walk based embedding method is performed to learn embeddings for each persona. Therefore it can obtain one or multiple representations for each node. However it blindly trains each persona to be close to the representation of its original node and cannot be trained in an end-to-end fashion. MCNE~\cite{wang2019mcne} is a GCN based method that utilizes binary mask layers to create multiple conditional embeddings for each node. MCNE is a supervised method and requires predefined aspects(ie. different user behaviors). MNE~\cite{yang2018multi} uses matrix factorization to obtain multiple embeddings for each node and considers the diversity of learned multiple embeddings. However, it ignores that the aspect of each node is dependent to its local structure. Park et al.~\cite{park2020unsupervised} proposes Asp2vec that determines the activated aspect of source node based on its context in current random walks. While the truncated walks cannot fully recover the neighborhood structure of nodes thus may provide a biased evidence on inferring node aspects, it is more desirable to model the multiple aspect of nodes via the detailed formation process of nodes.



\section{Conclusion}
In this paper, we present a novel multi-aspect embedding method called MHNE that models the aspect driven edge formation process of temporal networks via Mixture of Hawkes process. Moreover, we utilize Gumbel-Softmax with trainable temperature parameter to compute aspect distributions with different shapes. To better capture the closeness between source and history target nodes, graph attention mechanism is incorporated to assign larger weights for more convincing excitation effects. Experiments on eight real-world datasets demonstrate the effectiveness of MHNE.

\bibliographystyle{ACM-Reference-Format}
\bibliography{related_works_v2}

 \clearpage
\begin{appendices}
\setcounter{table}{0}

\section{Pseudocode}
\begin{algorithm}
    \caption{Pseudo code for MHNE}
    \LinesNumbered
    \textbf{Input}
    Temporal edges $\mathcal{E}=\{(u, v, t)|u\in\mathcal{V}, v\in\mathcal{V}\}$

    \textbf{Output}
    embedding matrix $\mathbf{I}\in\mathbb{R}^{\left| \mathcal{V} \right| *d}, \mathbf{A}\in\mathbb{R}^{\left| \mathcal{V} \right| *k*d}$
    
    \textbf{Initialize} embedding matrix $\{\mathbf{I}, \mathbf{A}\}$, and model parameters $\mathbf{\delta}, \mathbf{\tau}\in\mathbb{R}^{\left|\mathcal{V}\right|}, \mathbf{W} \in \mathbb{R}^{d*d}, \vec a \in \mathbb{R}^{(2d)*1}$
    
    Obtain history sequences $\mathcal{H}$ for each node

    \SetKwFunction{func}{{\rm lambda}}
    \SetKwProg{Fn}{Function}{:}{}

    \For{each edge $(u, v, t) \in \mathcal{E}$}{
        
        $C_u^k(t) = \frac{1}{2}\left(\frac{\sum_{h\in\mathcal{H}_u(t)}\mathcal{J}(t-t_h)\mathbf{A}_h^k}{|\mathcal{H}_u(t)|}+\mathbf{A}_u^k\right)$

        \For{each node $n\in\mathcal{H}_{u}(t_{v}) \cup \{u\}$}{
            
            $\pi_n^k(t) = \frac{\exp(\mathcal{F}(\mathbf{I}_n, C_u^k(t))+g_k)/\tau_n)}{\sum_{k^{'}}\exp(\mathcal{F}(\mathbf{I}_n, C_u^{k^{'}}(t))+g_{k^{'}})/\tau_n)}$
        }
        
        $\tilde \lambda_{v|u}(t)=$\func{$u, v, \mathcal{H}_u(t), \pi_n^k(t)$}
        
        Sample negative nodes $\mathcal{N}=\{v^i|(u, v^i, t) \notin \mathcal{E}\}$

        \For{each node $v^i \in \mathcal{N}$}{
            
            $\tilde \lambda_{v^i|u}(t)=$\func{$u, v^i, \mathcal{H}_u(t), \pi_n^k(t)$}
        }
        
        $\mathcal{L}=\log \sigma(\tilde{\lambda}_{v|u})+\sum_{i=1}^N\mathbb{E}_{v^i\sim p_n(v)}[-\log \sigma(\tilde{\lambda}_{v^i|u}(t))]$
        
        Update embedding matrices by maxmizing $\mathcal{L}$
    }

    \Fn{\func{$u, v, \mathcal{H}_u(t), \pi_n^k(t)$}}{
        
        $\mu_{u, v}=\mathcal{F}(\mathbf{I}_u, \mathbf{I}_v)$
        
        $attn_{u, h} = \frac{\exp(\mathbf{LeakyReLU}(\vec{a}^T(\mathbf{W}\mathbf{I}_u|\mathbf{W}\mathbf{I}_h)))}{\sum_{h^{'}}\exp(\mathbf{LeakyReLU}(\vec{a}^T(\mathbf{W}\mathbf{I}_u|\mathbf{W}\mathbf{I}_{h^{'}})))}$
        
        $\alpha_{h, v} = attn_{u, h} * \mathcal{F}(\mathbf{I}_h, \mathbf{I}_v)$

        \For{each aspect $k$}{
            
            $\gamma_{u, v}^k = -\mathcal{F}(\mathbf{A}_u^k, \mathbf{A}_v^k)$
            
            $\gamma_{h, v}^k = -\mathcal{F}(\mathbf{A}_h^k, \mathbf{A}_v^k)$
            
            $\tilde{\lambda}_{v|u}^k(t) = \mu_{u,v}\gamma_{u,v}^k+\sum_{h\in\mathcal{H}_u(t)}\pi_h^k\alpha_{h,v}\gamma_{h,v}^k\mathcal{K}(t-t_h)$
        }

    \textbf{Return} $\tilde{\lambda}_{v|u}(t) = \sum_{k}\pi_u^k \tilde{\lambda}_{v|u}^k(t)$
    }
    \textbf{EndFunction}
\end{algorithm}

\section{Dataset \& code}
\renewcommand\thetable{B. \arabic{table}}

\begin{table}[h]
\small
\caption{Datasets URL}
\begin{tabular}{@{}c|c@{}}
\toprule
Dataset  & URL                                                     \\ \midrule
DBLP     & https://dblp.uni-trier.de                               \\
Wosn     & http://networkrepository.com/fb-wosn-friends.php        \\
Digg     & http://networkrepository.com/digg-friends.php           \\
Epinions & http://networkrepository.com/soc-epinions-trust-dir.php \\
Tech     & http://networkrepository.com/tech-as-topology.php       \\
HepTh    & http://networkrepository.com/ca-cit-HepTh.php           \\
BtcAlpha & http://networkrepository.com/soc-sign-bitcoinalpha.php  \\
BtcOtc   & http://networkrepository.com/soc-sign-bitcoinotc.php    \\ \bottomrule
\end{tabular}
\label{Tab: Dataset url}
\end{table}

\begin{table}[h]
\caption{Code repositories}
\small
\begin{tabular}{@{}c|c@{}}
\toprule
Code     & URL                                        \\ \midrule
DeepWalk & https://github.com/phanein/deepwalk        \\
LINE     & https://github.com/tangjianpku/LINE        \\
asp2vec  & https://github.com/pcy1302/asp2vec         \\
M$^2$DNE    & https://github.com/rootlu/MMDNE            \\
HTNE     & https://zuoyuan.github.io/publication/HTNE \\ \bottomrule
\end{tabular}
\label{Tab: Code url}
\end{table}

Our method is implemented in PyTorch and the source code can be downloaded from: https://bit.ly/2LETW1W. Table.~\ref{Tab: Dataset url} shows the url links of used datasets. Table.~\ref{Tab: Code url} shows the repositories where the implementations of baseline methods can be downloaded. Nodes in all networks except for Tech refer to person with different personas~({i.e.} researcher), while nodes in Tech denotes autonomous systems. Specifically, Epinions, HepTh, BtcAlpha and BtcOtc are directed graphs while others are undirected. 

\section{Detailed Experiment settings}

For LINE, we set the number of total edge samples to be 1 billion, other parameters are set by default. We run LINE for two times with the order parameter set to be 1 and 2 respectively, representing the first and second-order proximity. After which we concatenate the learned 2 embeddings to obtain the final embedding for each node.

For DeepWalk and asp2vec, we set the number of walks, walk length, window size and number of negative samples to be 10, 80, 3 and 2 respectively as reported in ~\cite{park2020unsupervised}. For asp2vec, we set $\tau$ and $\lambda$ to be 0.5 and 0.01. Specifically, we set the threshold $\epsilon$ for aspect regularizer in asp2vec to be $0.9$.

For M$^2$DNE, HTNE and our method, we set the history length, number of negative samples both to be 5. We set batch size to be 1000 for all datasets except for BtcAlpha and BtcOtc, where we set batch size to be 200. Specifically, we set $\epsilon=0.4$ for M$^2$DNE.

Note that we train asp2vec, M$^2$DME, HTNE and MHNE using Adam with the same learning rate of 0.003 for 20 epochs. And we evaluate the learned embeddings from each epoch on downstream tasks~({i.e.} link prediction, temporal node recommendation) and report the best performance among them. We emperically find that asp2vec converges slowly than other methods on BtcAlpha and BtcOtc, thus we set the total training epochs for asp2vec to be 50 on these two datasets. 
    
For fairness, we set the parameters used for representing nodes to be the same for all methods. Which is if we set the embedding size $d=100$ for single embedding methods like HTNE, we set the dimension for each aspect embedding to be $100/(k+1)$, $k$ is the number of aspects, for MHNE and asp2vec obtains another identity(target) embedding for each node. And we obtain the final embedding which is fed into downstream task by concatenating all the learned embeddings for each node.
\end{appendices}

\end{document}